\documentclass[journal,twoside,web]{ieeecolor}
\let\labelindent\relax
\usepackage{tmi}
\usepackage{cite}
\usepackage{amsmath,amssymb,amsfonts}
\usepackage{algorithmic}
\usepackage{graphicx}
\usepackage{textcomp}
\usepackage{multirow}
\usepackage{enumitem} 

\def\BibTeX{{\rm B\kern-.05em{\sc i\kern-.025em b}\kern-.08em
    T\kern-.1667em\lower.7ex\hbox{E}\kern-.125emX}}
\begin{document}
\title{Towards Universal Text-driven CT Image Segmentation}
\author{Yuheng Li, Yuxiang Lai, Maria Thor, Deborah Marshall, Zachary Buchwald, David S. Yu and Xiaofeng Yang, \IEEEmembership{Member, IEEE}
\thanks{This research is supported in part by the National Institutes of Health under Award Number R01CA272991, R01DE033512, R01EB032680, R56EB033332 and U54CA274513.}
\thanks{Yuheng Li and Xiaofeng Yang, are with the Department of Biomedical Engineering, Georgia Institute of Technology, Emory University, Atlanta, GA 30332 USA.}
\thanks{Maria Thor is with the Department of Medical Physics, Memorial Sloan Kettering Cancer Center, New York, NY 10065.}
\thanks{Deborah Marshall is with the Department of Radiation Oncology, Icahn School of Medicine at Mount Sinai, New York, NY 10029.}
\thanks{Zachary Buchwald, David Yu and Xiaofeng Yang are with the Department of Radiation Oncology, Emory University School of Medicine, GA 30322 USA. (Corresponding author Xiaofeng YangXiaofeng Yang, e-mail: xiaofeng.yang@emory.edu).}}

\maketitle

\begin{abstract}
Computed tomography (CT) is extensively used for accurate visualization and segmentation of organs and lesions. While deep learning models such as convolutional neural networks (CNNs) and vision transformers (ViTs) have significantly improved CT image analysis, their performance often declines when applied to diverse, real-world clinical data. Although foundation models offer a broader and more adaptable solution, their potential is limited due to the challenge of obtaining large-scale, voxel-level annotations for medical images. In response to these challenges, prompting-based models using visual or text prompts have emerged. Visual-prompting methods, such as the Segment Anything Model (SAM), still require significant manual input and can introduce ambiguity when applied to clinical scenarios. Instead, foundation models that use text prompts offer a more versatile and clinically relevant approach. Notably, current text-prompt models, such as the CLIP-Driven Universal Model, are limited to text prompts already encountered during training and struggle to process the complex and diverse scenarios of real-world clinical applications. Instead of fine-tuning models trained from natural imaging, we propose OpenVocabCT, a vision-language model pretrained on large-scale 3D CT images for universal text-driven segmentation. Using the large-scale CT-RATE dataset, we decompose the diagnostic reports into fine-grained, organ-level descriptions using large language models for multi-granular contrastive learning. We evaluate our OpenVocabCT on downstream segmentation tasks across nine public datasets for organ and tumor segmentation, demonstrating the superior performance of our model compared to existing methods. All code, datasets, and models will be publicly released at https://github.com/ricklisz/OpenVocabCT.
\end{abstract}


\section{Introduction}
\label{sec:introduction}
\IEEEPARstart{I}{n} clinical practice, computed tomography (CT) is widely used for detailed anatomical visualization and precise segmentation of organs-at-risk and tumor lesions. With the advent of deep learning models such as convolutional neural networks (CNNs) \cite{ronneberger2015u, isensee2021nnu, roy2023mednext} and vision transformers (ViTs) \cite{tang2022self, zhou2023nnformer}, medical image segmentation has shown significant promise in CT image analysis. While many purpose-built models achieve remarkable precision for specific tasks \cite{lee20223d, zhao2022prior, marcus2023concurrent, ji2021learning}, their effectiveness is often limited when dealing with the generalizability observed in diverse multimodal clinical data or handling various imaging tasks.

\begin{figure*}[t]
	\centering
\includegraphics[width=\linewidth]{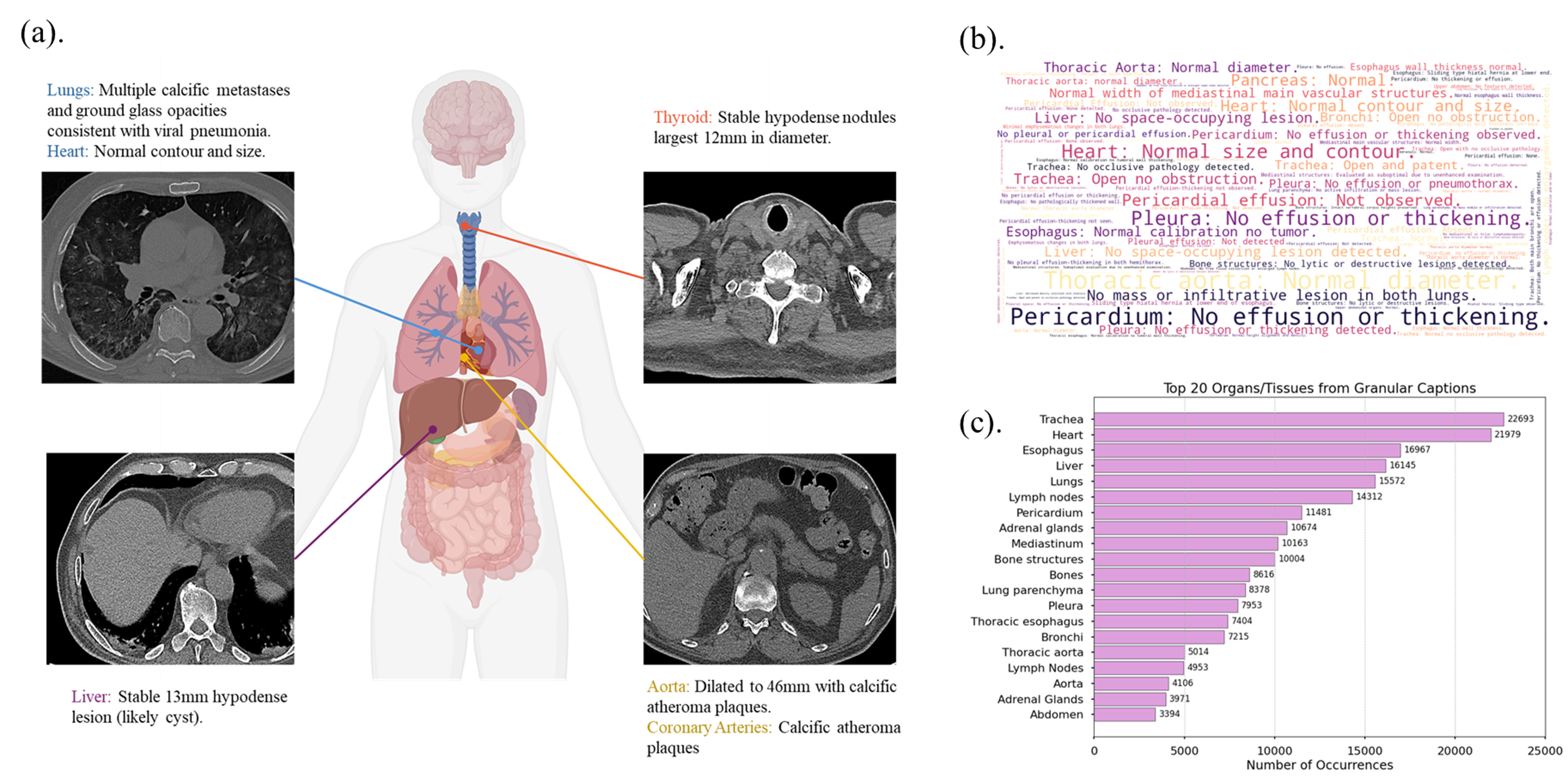}
	\caption{We use CT-RATE \cite{hamamci2024developing}, a large-scale paired CT and radiology report dataset, to generate pre-training data. Some key information about this dataset is present: (a) Example of paired CT scans and detailed captions for each organ; (b) Top 100 captions containing detailed organ-level information; (c) Distribution of the top 20 organs/tissues from these captions. We leverage pre-trained large language models to break down radiology reports into fine-grained captions and filtered out low-quality captions using substring matching for open-vocabulary segmentation training.}  \label{fig_data}
\end{figure*}

Foundation models, on the other hand, are designed to provide a broad base of capabilities that can be adapted to many tasks without extensive retraining \cite{moor2023foundation}. With the substantial increase in the availability of annotated imaging datasets, numerous studies have been conducted on building foundation segmentation models for 3D CT images \cite{silva2023towards, huang2023stu, wang2023mis}. However, obtaining abundant voxel-level annotations for medical images remains time-consuming and expensive. As a result, prompting-based foundation models are being explored as a more data-efficient solution \cite{huang2024segment, zhao2023one, koleilat2024medclip}. These models leverage either visual prompts (points or bounding boxes) or text prompts (natural language) to guide the segmentation process. While visual-prompting methods such as the Segment Anything Model (SAM) \cite{kirillov2023segment} are widely used for various tasks, they still face challenges when applied to medical settings due to: 1) SAM requires significant image-level annotations for training and considerable manual prompting for inference, which is not a scalable solution for clinical settings; 2) visual prompts lack the necessary clinical details to guide accurate segmentation. These limitations reduce the clinical utility of visual-prompting methods and underscore the importance of vision-language models, as both visual and textual features are essential for effective segmentation to facilitate the diagnosis, treatment planning, and treatment response assessment.

Text-prompting models have emerged as a promising solution for building foundation segmentation models in CT \cite{liu2023clip, zhao2023one}. Recent developments in learning visual representations from text supervision have shown tremendous success in computer vision \cite{radford2021learning} and medical imaging \cite{huang2021gloria}. However, prior works primarily focus on 2D medical images, such as chest X-rays \cite{wu2023medklip, wang2022medclip}, which limits their application to the more widely used volumetric CT imaging for individual segments of the whole-body. Additionally, text-driven segmentation methods for 3D CT scans often rely on text embeddings from pretrained text encoders, with limited or insufficient use of image-text alignment strategies \cite{li2023lvit, liu2023clip, zhao2023one}. This insufficient alignment can restrict the model’s ability to generalize effectively to diverse unseen clinical text prompts or varying medical vocabularies, hindering its utility in real-world clinical settings.

In real-world practice, healthcare professionals use varied prompts to describe or annotate specific organs, making generalization particularly important for achieving universal organ segmentation. Furthermore, radiology reports frequently contain detailed but lengthy diagnostic information, qualitatively deduced from the images, some of which is unrelated to organs or lesions, which poses challenges for efficient image-to-text alignment. To address these challenges, we developed a dedicated image-text supervision framework for 3D CT that generalizes to diverse and unseen text prompts, and that can be applied to text-prompt segmentation tasks. Our main contributions are summarized as follows:

1). We generated detailed organ-level image-text pairs from CT radiology reports by applying large language models (LLMs) on the large-scale paired CT and radiology report dataset, CT-RATE (n = 50,188) \cite{hamamci2024developing}.

2). We propose a pretraining framework that leverages organ-level and report-level textual information to align the vision-language model using a multi-label contrastive loss, outperforming other existing methods. 

3). We evaluate method on eight public datasets for organ and tumor segmentation, demonstrating its robustness across various text prompts and multi-target segmentation tasks. Compared to vision-only segmentation models, our approach achieves comparable performance while enhancing usability by allowing clinicians to interact with the model using natural language. Moreover, compared to text-driven models, it delivers superior performance in single-target segmentation tasks and generalizes effectively to diverse text prompts.


\section{Related Work}

\subsection{Medical image segmentation}
Deep learning-based methods \cite{ronneberger2015u, isensee2021nnu} have been widely applied for organ segmentation and tumor segmentation and detection, yielding promising results. However, these methods are often task-specific or organ-specific, such as organ segmentation \cite{pan2023abdomen} or tumor detection \cite{xue2021multi,chen2021effective}. Recently, there has been a growing interest in building foundation segmentation models for various organs and tumors \cite{huang2023stu, liu2023clip, xie2022unimiss}, and for this reason adapting SAM for 'universal' segmentation \cite{huang2024segment,ma2024segment}; however, SAM’s interactive nature still requires considerable manual input, limiting its applicability in clinical settings \cite{shi2023generalist}. Additionally, since SAM was developed using natural images, it may lack the medical semantic understanding to differentiate between healthy organs and tumors. In contrast, our method integrates medical professional's knowledge from radiology reports into a text model, which is more practical than visual prompting for clinical usage.

\subsection{Vision-language model for medical imaging}
Vision-language models such as CLIP \cite{radford2021learning} have demonstrated the ability to learn transferable visual features through language supervision, without manual image annotations. In medical imaging, where diagnostic radiology-founded reports complement imaging data, CLIP-based models have shown promise across various tasks, including organ segmentation \cite{liu2023clip}, disease classification \cite{wu2023medklip}, and image-text retrieval \cite{zhang2023biomedclip}. However, adapting CLIP to medical imaging presents unique challenges. Unlike natural images, which can often be summarized in a few sentences, medical images like CT scans contain complex diagnostic information, resulting in detailed radiology reports. This complexity demands improved local-level alignment between image and text. Another significant challenge is data scarcity. While CLIP is ideal for large-scale datasets, medical image-text datasets are relatively limited, which requires a more label-efficient approach. Our method addresses this by curating a radiology-specific corpus of image-text pairs, designed to supplement the radiology-specific knowledge that conventional CLIP training lacks.

\subsection{Text-driven segmentation model}
Prompt engineering has demonstrated strong performance improvements in both natural language processing and computer vision tasks. By utilizing pre-trained vision-language models \cite{radford2021learning}, text prompting methods results in open-vocabulary segmentation \cite{liang2023open, ghiasi2022scaling} and referring segmentation \cite{wang2022cris} tasks. Previous studies demonstrate that pretrained CLIP models are effective for 2D medical image segmentation tasks \cite{muller2022radiological, koleilat2024medclip}. Li et al. propose a text-augmented segmentation model that utilizes medical text annotations and pseudo labels in a semi-supervised framework \cite{li2023lvit}. However, extending these approaches to 3D vision-language models for CT segmentation remains an active research area due to the limited availability of paired CT image-text datasets. Recent efforts have centered on developing text-prompted universal models for segmenting various organs and tumors in 3D volumes. Liu et al. proposed to leverage CLIP's text embeddings to guide the segmentation model for partially-labeled datasets \cite{liu2023clip}. Zhao et al. introduced SAT, a large-scale segmentation model with a knowledge-enhanced text encoder for multimodal organ and tumor segmentation \cite{zhao2023one}. However, no existing approach fully leverages a vision-language-aligned model like CLIP for 3D medical image segmentation tasks. Given CLIP’s strong grounding capabilities, we argue that vision-language alignment is essential for building robust text-driven segmentation models. To address this, we propose a pre-trained vision-language model trained on a large-scale CT image-report dataset, incorporating diverse captions for each organ and region.
\section{Method}

\begin{figure*}[t]
	\centering
\includegraphics[width=0.8\linewidth]{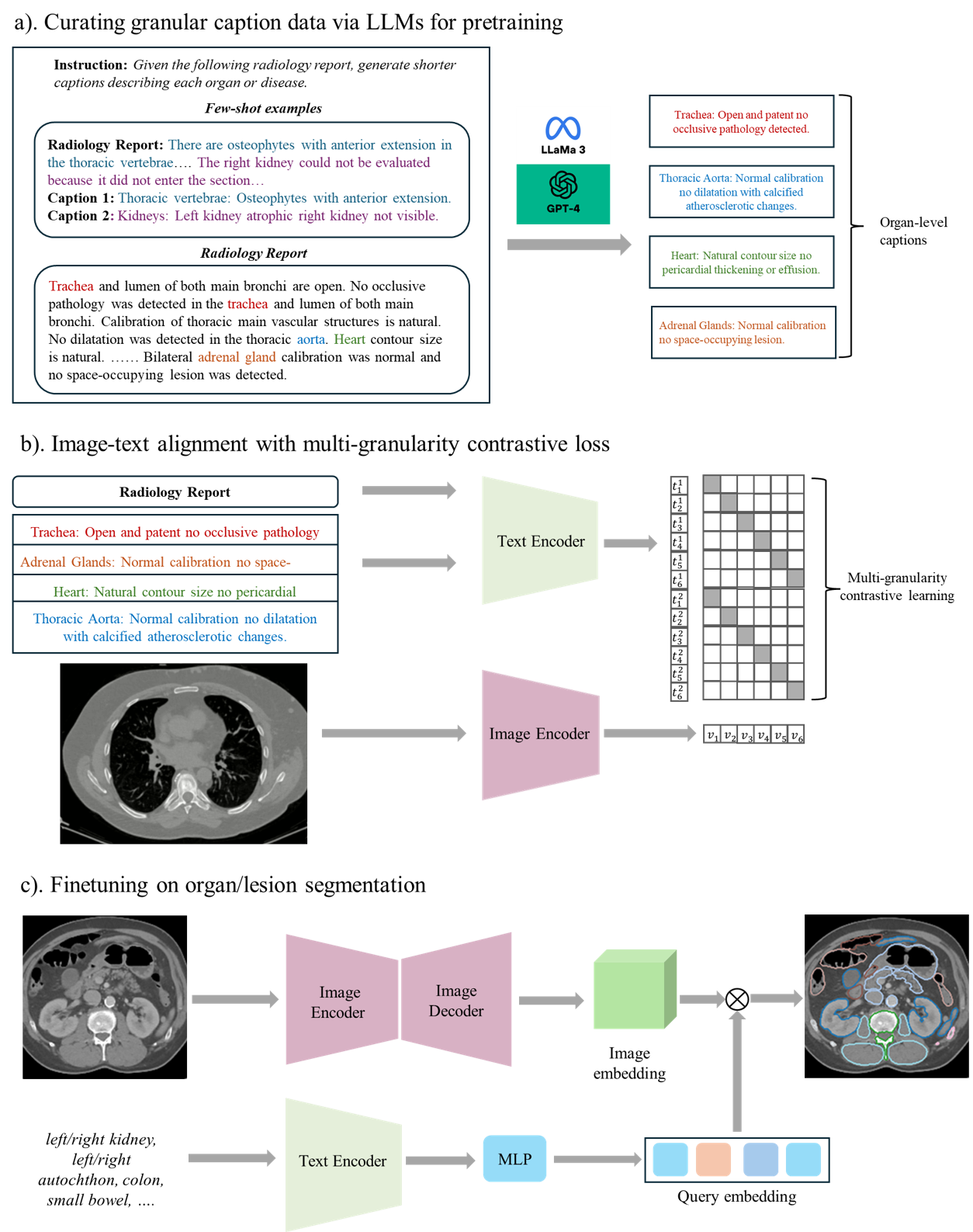}
	\caption{Overall workflow for OpenVocabCT. (a) We first curate granular organ-level captions from CT-RATE’s radiology report using LLMs with few-shot examples. The LLMs break down long radiology findings into organ-level captions, which are further filtered via substring matching to our metadata. (b) We pretrain our vision language model using a multi-granularity contrastive loss. Each CT image is paired with multiple granular captions and the original report to enhance text representation learning. (c) We finetune the vision language model on CT segmentation datasets with text prompts for each organ.}  \label{fig_pipeline}
\end{figure*}

\subsection{Preliminary}
In recent years, language supervision methods such as CLIP \cite{radford2021learning} and SigLIP \cite{zhai2023sigmoid} have been shown to be effective in learning transferable representations. Concretely, given a batch of images \{$I_1$,$I_2$,···,$I_N$\} and paired text descriptions \{$T_1$,$T_2$,···,$T_N$\}, CLIP leverages an image encoder $E_i$ to extract the image embedding $v_N$ and a text encoder $E_t$ to extract the text embedding $t_N$:
\begin{equation}v_N = E_i(I_N;\theta),t_N = E_t(T_N;\beta).\label{eq}\end{equation}
The extracted image embedding and text embedding are used to compute InfoNCE loss, in which paired image-text samples are considered as positives and the unpaired ones as negatives. The image-to-text loss ${L}_{i2t}$ can be formulated as:
\begin{equation} {L}^{i2t}_{\text{CLIP}} = -\sum_{i=1}^{N} \log \left( \frac{\exp(\text{cos}(v_i, t_i) / \tau)}{\sum_{j=1}^{N} \exp(\text{cos}(v_i, t_j) / \tau)} \right) .\label{eq}\end{equation}
where cos⟨·,·⟩ denotes the cosine similarity and ${\tau}$ is a learnable temperature parameter. The final bidirectional total loss can be formulated as $\mathcal{L}_{\text{CLIP}} = \frac{1}{2} (\mathcal{L}^{i2t}_{\text{CLIP}} + \mathcal{L}^{t2i}_{\text{CLIP}})$. 
\subsection{Improving image-text alignment with language models}
While CLIP has demonstrated promising results on natural images using large-scale datasets, it still faces considerable challenges in medical data due to the significant domain gap between them. First, paired medical image and text data are much more limited compared to natural images, necessitating the use of image-level and text-level data augmentations. Second, medical text typically comes in the form of diagnostic reports made by radiologists, which are much longer and more complex than the concise captions found in natural image datasets. While diagnostic reports contain abundant information about the patient, they may lack the granular, organ-level details essential for effective text-driven segmentation. Also, directly training the model with these lengthy reports may cause it to overgeneralize, attempting to encompass all organs and abnormalities in a single image without distinguishing among them.

Recognizing these challenges, we propose to leverage existing LLMs to generate granular text descriptions relating to organs and abnormalities from each CT image's radiology report. As in-context learning shows great promise in aligning LLM's responses to human gold standards, we also leverage this technique to generate organ-level descriptions. First, we generate some few-shot examples of concise organ-level captions from the radiology impressions (Figure 1). We then prompt GPT-4 and Llama-3 APIs with "\textit{You are a medical expert. You will read radiology reports from various physicians on CT images. Given the following radiology report, generate shorter captions describing each organ or disease.}" to generate target texts. The LLMs will receive the few-shot examples as context and the radiology report as input data, returning the generated organ-level descriptions. By using in-context learning, we ensure the model maintains consistency in its response and preserves the semantic details in the original caption. Finally, observing that the generated captions may contain redundant information, we followed MetaCLIP's \cite{xu2023demystifying} approach to filter out low-quality captions via substring matching. We utilize a radiology-specific text corpus RadLex, which provides an agreed set of named entities for radiology procedures, as our text metadata. We then apply sub-string matching on the granular captions with the metadata entries, which identifies high-quality captions that contain any of the metadata entries, filtering the various types of noise that the LLM may introduce. By addressing these challenges, we aim to enhance the training process and expand the benefits of augmentation to the text inputs, leading to improved performance and a more comprehensive learning framework.

\subsection{Multi-granularity contrastive learning}
Having generated and filtered the image captions, we can improve the image-text alignment via a multi-granularity contrastive learning framework. The main difference between our approach and CLIP's is that our approach leverages multiple granular captions as text augmentation. Given the granular captions $C^l = [c_1, \dots, c_M]$, we implement a simple random sampling strategy to gather short captions:
\begin{equation}
S_{i,j} \sim \text{Uniform}([c_1, \dots, c_M]),
\end{equation}
where $S_{i,j}$ refers to the $j$-th caption of $i$-th sample from the short caption set. The multi-granular contrastive text-to-image loss becomes:
\begin{equation}
\mathcal{L}^{t2v}_{\text{MGCL}} = - \sum_{i=1}^{N} \sum_{j=1}^{K} \log \frac{\exp\left( \text{cos}(\mathbf{v}_i, \mathbf{t}_{i,j})/\tau \right)}{\sum_{n=1}^{N} \exp \left( \text{cos}(\mathbf{v}_n, \mathbf{t}_{i,j})/\tau \right)},
\end{equation}
where $\mathbf{t}_{i,j}$ refers to the text embedding of $S_{i,j}$, and $K$ denotes the number of randomly sampled short captions (K=3 in our implementation). We formulate the bi-directional multi-granular contrastive loss: $\mathcal{L}_{\text{MGCL}} = \left(\mathcal{L}^{t2v}_{\text{MGCL}} + \mathcal{L}^{v2t}_{\text{MGCL}}\right)/2$. 
By introducing short captions as text augmentations, we increase the diversity of the training data, resulting in a more complete and aligned representation of both images and text. 

Empirically, we find that image-text pretraining may not result in optimal finetuning performance for dense prediction tasks like segmentation. To enhance the model's dense prediction capability, we first pretrained the image encoder on TotalSegmentator for segmentation for 500 epochs, providing a good initialization for the image encoder. We then initialized the image encoder with the pretrained model's weights before image-text alignment. During image-text contrastive learning, we locked the image encoder and only tuned the text encoder to align the text embeddings with the vision embeddings.

For the final pretraining objective, we combine the original CLIP loss $\mathcal{L}_{\text{CLIP}}$ using the full radiology report with the proposed $\mathcal{L}_{\text{MGCL}}$ loss using the generated short captions: 
\begin{equation}
\mathcal{L}_{\text{Final}} = {L}_{\text{CLIP}} + {L}_{\text{MGCL}},
\end{equation}
This approach helps the model to capture detailed connections between images and text while keeping the broader context provided in the radiology reports, improving its ability to generalize to a variety of image-text pairs. 

\section{Experiments}
\subsection{Datasets and evaluation metrics}
\subsubsection{Vision-language pretraining dataset}{We leverage a large-scale chest CT dataset CT-RATE \cite{hamamci2024developing}, that have 21,304 patients and 50,188 image-radiology report pairs. The CT images were obtained using a range of reconstruction techniques. The radiology reports have four sections: 1. clinical information inlcuding symptoms and history, 2. imaging technique and acquisition protocol, 3. imaging findings (anatomical/pathological observations), and 4. impression/diagnosis.}
\subsubsection{Segmentation datasets}{For finetuning on segmentation datasets, we connect our pretrained encoder to STUNet decoder and finetuned on a variety of datasets: TotalSegmentator \cite{wasserthal2023totalsegmentator}, MSD Lung, MSD Pancreas, MSD Hepatic Vessel, MSD Colon, MSD Liver \cite{antonelli2022medical}, KiTS23 \cite{heller2023kits21}. For each dataset, we used 80\% for training, and 20\% for testing. We also evaluate the segmentation model's generalization capabilities on the FLARE22 and SegTHOR datasets. Dice Similarity Coefficient (DSC) is utilized to quantitatively evaluate organ/tumor segmentation performance.}

\subsection{Implementation details}
\subsubsection{Pretraining}{Our OpenVocabCT is composed of an image encoder, a text encoder and a segmentation decoder. For pretraining, we use the STUNet-Large \cite{huang2023stu} as the backbone for our image encoder due to its excellent performance in various benchmarks. We use BIOLORD \cite{remy2024biolord} as our text encoder, which was pretrained on both clinical sentences and biomedical concepts. The text encoder's latent feature is projected to the image encoder's latent feature dimension using a simple MLP. We preprocessed CT images by resampling them to an isotropic spacing of 1.5 mm × 1.5 mm × 1.5 mm and padding them to a size of 220 × 220 × 220. For image captions, we randomly sample three granular captions from the filtered dataset, along with the findings section from the original report. Pretraining is conducted on four NVIDIA A100 GPUs, with a batch size of 32 per GPU. \\
(2). Finetuning on segmentation datasets: We directly use the aligned image encoder and text encoder from pretraining. We then connect the pretrained image encoder to the decoder of STUNet-Large. To avoid catastrophic forgetting, we freeze the weights of both image encoder and text encoder and only tune the segmentation decoder. For text-driven segmentation, we generate organ prompts such as 'Liver' or 'Left kidney' and feed the tokenized prompts to the text encoder similar to \cite{liu2023clip, zhao2023one}. The text features are further processed through a text-guidance connector to generate query embeddings, which are then multiplied with the image features to produce segmentation maps. In our ablation study, we explored various strategies for designing the text-guidance connector.}

\subsection{Comparison with other methods}
\begin{figure*}[t]
	\centering
\includegraphics[width=0.9\linewidth]{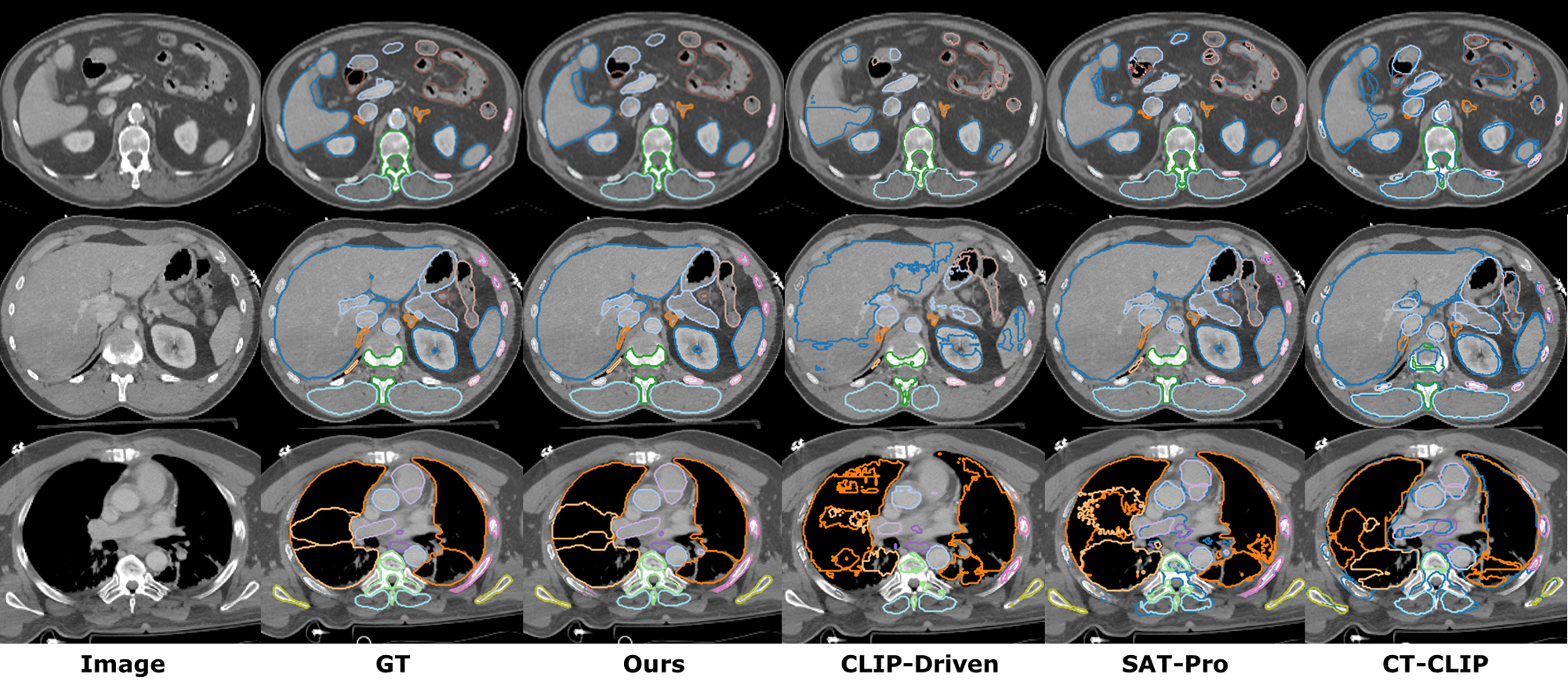}
	\caption{Visualization of organ segmentation on TotalSegmentator. Each segmentation model is evaluated under training seen prompts.}  \label{total_result}
\end{figure*}

{We compare our method with both vision-only segmentation models and text-driven segmentation frameworks. For vision-only methods, we compare with nnUNetv2 \cite{isensee2021nnu} and UniMiSS \cite{xie2022unimiss}. For text-driven approaches, we compare with CLIP-Driven Universal Model \cite{liu2023clip}, SAT-Pro \cite{zhao2023one}, and CT-CLIP \cite{hamamci2024developing}. As shown in Table~\ref{table1}, our method is able to achieve better performance than both the previous SOTA image-only methods and the text-driven SOTA methods on TotalSegmentator dataset. Compared to the best image-only method UniMISS, our method outperforms by 2.7\% DSC. Compared to the best text-driven method, i.e., SAT, our method outperforms by 3.1\% DSC. We also visualize the organ segmentation results in (Figure~\ref{total_result}). This shows that our method can effectively segment the majority of organs with superior performance. For tumor segmentation tasks, as shown in Table~\ref{table2}, our method achieves comparable performance as the best text-driven method Universal model and outperforms the best vision-only method nnUNet by 4.3\%  on average.}

\begin{table*}
\caption{Finetuning performances on TotalSegmentator. Results reported in DSC (\%). \textbf{BOLD} indicates best result. \underline{Underline} second best result.}
\label{table1}
\centering
\begin{tabular}{c|c|c|c|c|c|c}
\hline
Method & Vertebrae & Cardiac & Muscles & Organs & Ribs & Avg \\
\hline
CLIP-Driven & 81.1 & 84.5 & \underline{88.8} & 86.4 & 82.1 & 84.6 \\
SAT Pro          & 85.4 & \underline{89.2} & 88.0 & 87.7 & 83.7 & 87.6 \\
CT-CLIP          & 76.6 & 78.1 & 81.2 & 79.1 & 74.2 & 77.8 \\
UniMiSS          & 85.1 & 88.9 & 92.8 & 88.5 & 84.5 & \underline{88.0} \\
nnUNet           & \underline{87.0} & 88.7 & 85.1 & \underline{87.5} & 86.1 & 86.9 \\
\textbf{OpenVocabCT}    & \textbf{90.4} & \textbf{90.3} & \textbf{90.0} & \textbf{91.3} & \textbf{91.6} & \textbf{90.7} \\
\hline
\end{tabular}
\end{table*}

\begin{table*}[ht]
\centering
\caption{Finetuning performance on tumor Segmentation. Results are reported in DSC (\%) for different datasets (MSD Lung, Pancreas, Hepatic Vessel, Colon, Liver, and KiTS23) across various segmentation tasks. \textbf{BOLD} indicates best result. \underline{Underline} second best result.}
\label{table2}
\begin{tabular}{l|c|ccc|ccc|c}
\hline
\multirow{2}{*}{\textbf{Method}} & \textbf{MSD Lung} & \multicolumn{3}{c|}{\textbf{MSD Pancreas}} & \multicolumn{3}{c|}{\textbf{MSD Hepatic Vessel}} & \textbf{MSD Colon} \\
\cline{2-9}
& Lung Tumor & Pancreas & Pancreas Tumor & Avg & Hepatic Vessel & Hepatic Vessel Tumor & Avg & Colon Tumor \\
\hline
CLIP-Driven & 67.1 & \textbf{82.7} & \textbf{60.8} & \textbf{71.7} & 62.6 & 69.4 & 66.0 & \underline{62.1}\\
SAT Pro         & 61.8 & 76.2 & 41.6 & 58.9 & 65.2 & 61.8 & 63.5 & 32.4 \\
CT-CLIP         & 52.9 & 68.3 & 33.2 & 50.8 & 52.9 & 55.0 & 53.9 & 28.1 \\
nnUNet          & \underline{68.2} & 81.6 & 53.1 & 67.4 & \textbf{67.7} & \textbf{72.1} & \textbf{69.9} & 49.2 \\
\textbf{OpenVocabCT}   & \textbf{70.3} & \underline{81.9} & \underline{60.0} & \underline{71.0} & \underline{67.3} & \underline{70.2} & \underline{68.8} & \textbf{62.8} \\
\hline
\end{tabular}

\vspace{1em} 

\begin{tabular}{l|ccc|cccc|c}
\hline
\multirow{2}{*}{\textbf{Method}} & \multicolumn{3}{c|}{\textbf{MSD Liver}} & \multicolumn{4}{c|}{\textbf{KiTS23}} & \multirow{2}{*}{\textbf{Avg}} \\
\cline{2-8}
& Liver & Liver Tumor & Avg & Kidneys & Kidney Cysts & Kidney Tumor & Avg & \\
\hline
CLIP-Driven & \underline{96.5} & \textbf{71.9} & \textbf{84.2} & 95.2 & \underline{76.4} & \underline{84.7} & \underline{85.4} & \underline{75.4} \\
SAT Pro         & 92.7 & 59.7 & 76.2 & 93.2 & 52.8 & 68.2 & 71.4 & 64.1 \\
CT-CLIP         & 86.7 & 51.8 & 63.5 & 85.4 & 42.6 & 52.2 & 60.1 & 55.4 \\
nnUNet          & 93.8 & 66.0 & 79.9 & \underline{96.1} & 58.0 & 84.4 & 79.5 & 71.8 \\
\textbf{OpenVocabCT}   & \textbf{96.6} & \underline{68.5} & \underline{82.6} & \textbf{96.4} & \textbf{78.0} & \textbf{86.9} & \textbf{87.1} & \textbf{76.2} \\
\hline
\end{tabular}
\end{table*}

\subsection{Generalizability to diverse text prompts}
{Compared to vision-only models, text-driven segmentation models are more flexible by parsing a wide range of clinical descriptions to guide the segmentation process. This allows text-driven models to generalize to partially labeled data that are incomplete or inconsistent as compared to training data. In real-world scenarios where healthcare professionals use varied prompts to describe or annotate specific organs, this generalization capability becomes particularly valuable for achieving universal organ segmentation. To assess these generalization capabilities, we evaluate how well the text-driven segmentation model handles two categories of training invisible prompts: 1) prompts obtained by merging multiple organs and 2) prompts that are synonymous terms for the target organ. Specifically, for category 1, we obtain these prompts by merging various suborgans used in training (i.e. \textit{left lung} is obtained by merging the \textit{lung upper lobe left}, \textit{lung lower lobe left} classes in TotalSegmentator). For category 2, we take the training visible prompt and substitute it for a synonym. For example, renal organs is synonym for kidney and hepatic system is synonym for liver.  

Generalization results to merged suborgans is shown in Table~\ref{table3}. Compared to the CLIP-Driven Universal Model, SAT Pro, and CT-CLIP, our method consistently achieves superior performance on merging simple left and right organs (e.g. \textit{left and right lungs}, \textit{left and right kidneys}). These results highlight the model's ability to interpret novel combinations of suborgans effectively. Generalization results to synonyms is shown in Table~\ref{table4}. Notably, our method also achieves significantly higher performance in challenging cases (e.g., \textit{cervical vertebrae, lumbar vertebrae, thoracic vertebrae}, \textit{veins}, \textit{arteries}) without explicitly being trained on such prompts. Our method consistently outperforms the existing text-driven methods, achieving the highest average performance (73.2\% DSC) across all categories.

We visualize the segmentation results of generalization study in axial view (Figure~\ref{generalize_axial}) and coronal view (Figure~\ref{generalize_coronal}). As shown in Figure~\ref{generalize_axial} rows 1, 3, and 4, our model performs resonably well when merging left and right organs in both the chest region (lung) and the abdominal region (kidney, autochthon). In row 2, the model demonstrates its flexibility by accurately segmenting organs described using synonyms, such as "hepatic system" for the liver and "renal organs" for the kidneys. Additionally, for bones and vertebrae, our method effectively segments merged categories like left ribs, right ribs, lumbar vertebrae, and thoracic vertebrae (Figure~\ref{generalize_axial}). This demonstrates the superior ability of our model to handle diverse and unseen clinical terminology as text prompts, suitable for real-world deployment in diverse clinical environments.}

\begin{table*}
\caption{Generalizability to training invisible text prompts: merging suborgans. L.: Left. R.: Right. AG: Adrenal gland. V.: Vertebrae. \textbf{BOLD} indicates best result. \underline{Underline} second best result.}
\label{table3}
\centering
\begin{tabular}{c|c|c|c|c|c|c|c|c|c|c}
\hline
Method & 
L. Lung	& R. Lung	& L\&R Lung & L.Heart & R. Heart	& L. and R. Kidney &  L. and R. AG & Heart & L. Ribs	& R. Ribs\\
\hline
CLIP-Driven & \underline{80.8}	& 50.1	& 21.0	& 42.3 &	64.0	& 45.7	& \underline{40.4}  & \textbf{63.1} & 0	& 10.3\\
SAT Pro & 63.2	& \underline{62.4} & 6.0	& 0	& 52.9	& 28.7	& 34.2 & 8.4 & \underline{63.2} & \underline{62.4} \\
CT-CLIP &48.5	&47	&\underline{26.5}	&\underline{47.0}	&54.2	&\underline{64.9}	&0 &25.7	&3.1	&0\\
\textbf{OpenVocabCT} & \textbf{95.4}	& \textbf{92.8}	& \textbf{85.6}	& \textbf{51.3}	& \textbf{68.2}	& \textbf{89.3}	& \textbf{77.3} & \underline{54.2} & \textbf{68.1}	&\textbf{66.7}\\
\hline
\end{tabular}

\vspace{1em} 

\begin{tabular}{c|c|c|c|c|c|c|c|c|c}
\hline
Method
& Trachea and Esophgaus	& L. Gluteus	& R.Gluteus	& Cervical V.	& Lumbar V. & Thoracic V. & Veins	& Arteries & Avg\\
\hline
CLIP-Driven &\textbf{69.0}	&\underline{50.0}	&19.8	&20.3	&5	&\underline{21.1}	&36.1	&\underline{63.1} & \underline{39.0}\\
SAT Pro & 1.1	&49.6	&\underline{49.8}	&\underline{29}	&4.9	&13.8	&10.4	&5	&  23.3\\
CT-CLIP &47.1	&1.2	&5.6	&0	&\underline{22.8}	&0	&\underline{38.9} &52.8 & 26.9\\
\textbf{OpenVocabCT} &\underline{67.1}	&\textbf{78.1}	&\textbf{77.5}	&\textbf{66.2}	&\textbf{31.6}	&\textbf{43.0}	&\textbf{62.8}	&\textbf{63.3} & \textbf{68.8}\\
\hline
\end{tabular}
\end{table*}

\begin{table*}
\caption{Generalizability to training invisible text prompts: synonyms. Parentheses represent the ground truth class. \textbf{BOLD} indicates best result. \underline{Underline} second best result.}
\centering
\begin{tabular}{c|p{1.8cm}|p{1.9cm}|p{1.9cm}|p{1.9cm}|p{1.9cm}|p{1.9cm}|c}
\hline
\multirow{2}{*}{Method} & \centering Renal organs (kidney)	& \centering Hepatic system (liver)	& \centering Heart muscle (myocardium)		& \centering Aortic vessel (aorta)	& \centering Cerebrum (brain)	& \centering Small intestine (small bowel) & \multirow{2}{*}{Avg}\\
\hline
CLIP-Driven & \centering \underline{54.5}	& \centering 0	&\centering 37.4 & \centering 42.0	& \centering \underline{78.5} & \centering 24.3 & 39.4\\
SAT Pro & \centering 0	& \centering \underline{74.7}	& \centering \textbf{76.5}	& \centering \textbf{83.6}	& \centering 0	& \centering \underline{76.2} & \underline{51.8}\\
CT-CLIP &\centering23.6	&\centering45.7	&\centering65.1		&\centering57.9	&\centering72.5	&\centering19.3 & 47.4\\
\textbf{OpenVocabCT} &\centering \textbf{77.1}	&\centering \textbf{84.5}	& \centering \underline{69.0} &\centering \underline{71.1}	&\centering \textbf{79.8}	&\centering \textbf{79.9} & \textbf{76.9}\\
\hline
\end{tabular}
\label{table4}
\end{table*}

\begin{figure*}[t]
	\centering
\includegraphics[width=0.9\linewidth]{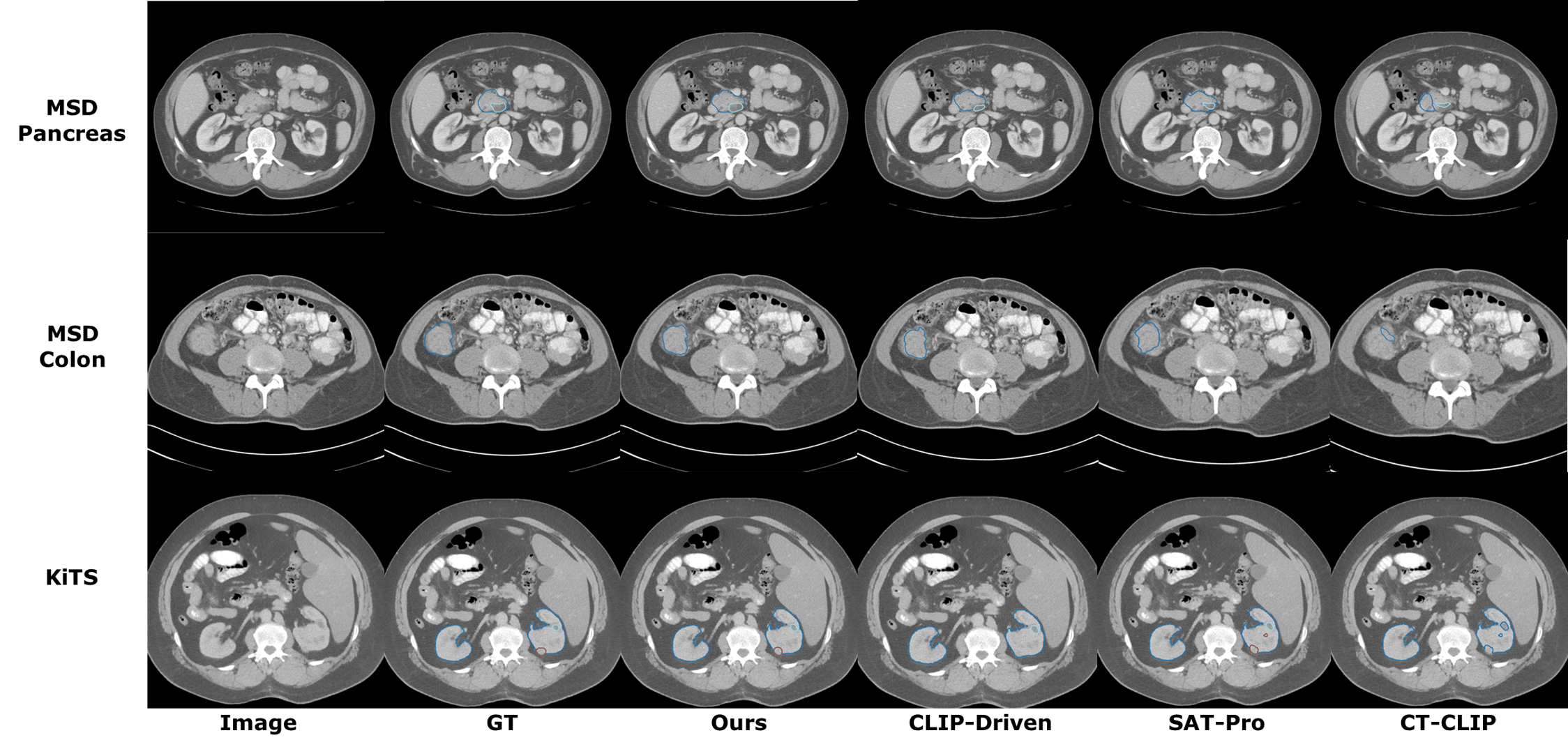}
	\caption{Visualization of tumor segmentation on MSD Pancreas, MSD Colon and KiTS. Each segmentation model is evaluated under training seen prompts.}  \label{msd_result}
\end{figure*}

\begin{figure*}[t]
	\centering
\includegraphics[width=0.9\linewidth]{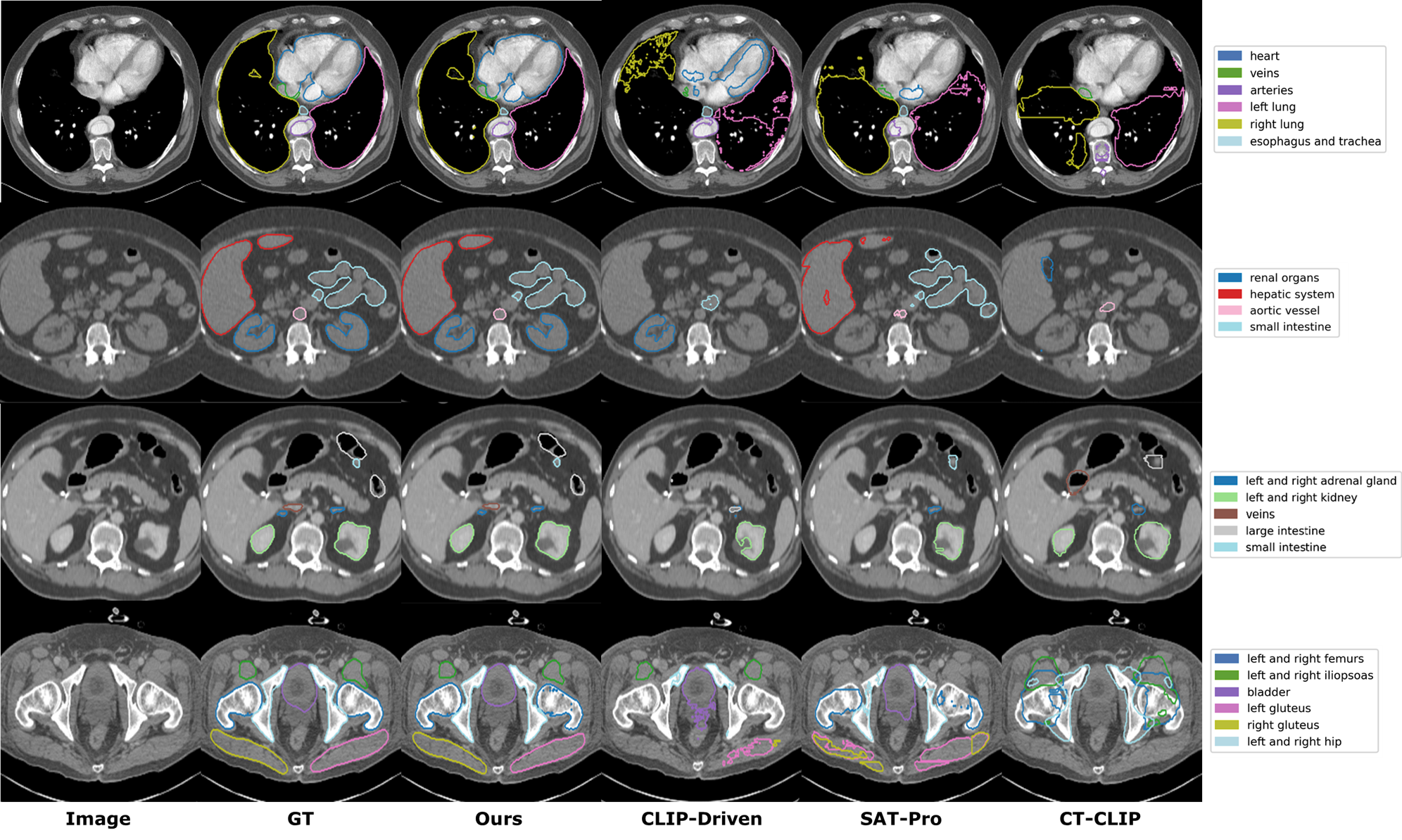}
	\caption{Visualization of generalization study (axial view). Each segmentation model is evaluated under training unseen prompts, as depicted in the corresponding color-coded legend.}  \label{generalize_axial}
\end{figure*}

\begin{figure*}[t]
	\centering
\includegraphics[width=\linewidth]{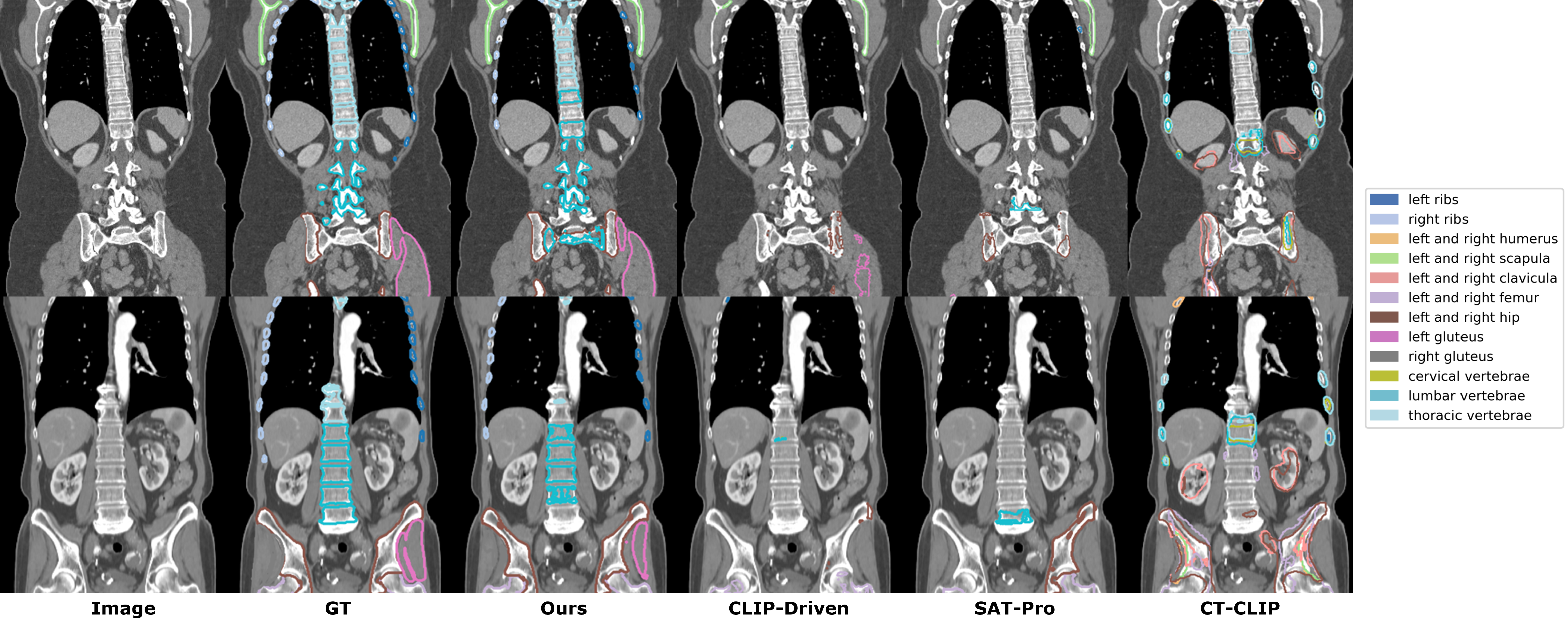}
	\caption{Visualization of generalization study (coronal view). Each segmentation model is evaluated under training unseen prompts, as depicted in the corresponding color-coded legend.}  \label{generalize_coronal}
\end{figure*}

\subsection{Ablation Study}
\subsubsection{Ablation study on pretraining}
{We conduct an ablation study on the effectiveness of our proposed pretraining strategy, shown in Table~\ref{table5_ablation}. Compared with a baseline model using random initialization, CLIP pretraining does not improve the finetuning performance, which corroborates our hypothesis that image-text alignment may not benefit a dense segmentation task. Incorporating our proposed MGCL on average improves the finetuning performance by 1.2\% DSC and the generazaibility performance by 6.8\% DSC and 12.4\% DSC. We find that the optimal performance emerges when initializing the image encoder with pretrained weights and locking the image encoder, considerably improving finetuning score by 8.5\% DSC and generalizability scores by 7.7\% DSC for merging suborgans.}

\subsubsection{Ablation study on text-branch connector}
{We also conduct an ablation study on 2 architectures of the text-branch connector: Multi-Layer Perceptron (MLP) vs Cross-Attention mechanism in Table~\ref{table5_ablation}. Intriguingly, we find that the MLP connector achieves strong generalizability (61.1\% DSC for merging and 73.1\% DSC for synonyms) by efficiently aligning image and text without overfitting. In contrast, while cross-attention can result in a higher finetuning score of 85.6\% DSC, it results in significantly lower generalizability, likely due to its reliance on specific text features. To mitigate this trade-off, our approach leveraging pretrained image encoder can provide a good initialization weight for our image-text alignment, achieving the best average performance, balancing the finetuing and generalizabitlity to diverse text prompts.}

\begin{table*}
\centering
\caption{Ablation Study on Pretraining Strategies and text-branch connector. Results are reported in DSC (\%).}
\label{table5_ablation}
\begin{tabular}{l|l|c|c|c}
\hline
\multirow{2}{*}{Pretraining Strategy} & \multirow{2}{*}{Segmentation text-branch} & \multirow{2}{*}{Finetuning} & \multicolumn{2}{c}{Generalizability} \\
\cline{4-5}
 &  & &Merging & Synonyms \\
\hline
Random Init & MLP & 83.4	& -  & - \\
CLIP Loss & MLP & 81.0	& 54.3 & 60.7 \\
Multi-Granularity Contrastive Loss & MLP & 82.2 & 61.1 & 73.1 \\
Multi-Granularity Contrastive Loss & Cross Attention & 85.6 & 43.4 & 49.4 \\
Pretrained Image Encoder + Multi-Granularity Contrastive Loss & MLP & 90.7 & 68.8 & 73.2 \\
\hline
\end{tabular}
\end{table*}

\subsection{Generalization study}
{We further study the generalization capabilities to external datasets. Table~\ref{tab:flare22_generalization} summarizes the performance on FLARE22 dataset. Compared to the other methods, our method consistently achieves superior generalization performances in 10 of 13 organs in the abdominal region. The second best performing model is the vision-only model nnUNet, which our method slightly outperforms by 0.4\% DSC. Generalization performance for SegTHOR dataset is shown in table~\ref{tab:segthor_generalization}. Our method also achieves superior performance in esophagus, heart and aorta segmentation. For inferring the training unseen heart category, we prompt text-driven models with the prompt \textit{heart}. Our method outperforms best existing text-driven models in heart segmentation by 9.4\% DSC and 6\% DSC on average. To further explore the generalization capability of nnUNet on the unseen heart category, we combined predictions for its sub-organ components (i.e., heart myocardium, heart ventricle, heart atrium, and pulmonary artery). Interestingly, nnUNet demonstrates comparable performance on this cardaic organ category, achieving results superior to most existing text-driven models (except ours). We hypothesize that this is because text-driven models may suffer from insufficient image-text alignment during training, limiting their ability to generalize effectively to unseen categories (heart in this case).}

\begin{table*}[ht]
\caption{Generalization study on the Flare22 dataset. All models are trained only on TotalSegmentator for fair comparison. Results are reported in DSC (\%). \textbf{BOLD} indicates best result. \underline{Underline} second best result.}
\centering
\renewcommand{\arraystretch}{1.2} 
\setlength{\tabcolsep}{4pt} 
\begin{tabular}{l|ccccccccccccc|c}
\hline
Method & Liver & RK & LK & LAG & RAG & Spleen & Pancreas & Gallbladder & Esophagus & Stomach & Duodenum & IVC & Aorta & \textbf{Avg} \\ \hline
Universal & 95.0 & 87.9 & 88.3 & 83.1 & 83.7 & 88.1 & 79.3 & 87.8 & 84.5 & 93.5 & 66.7 & 91.4 & 90.6 & 86.2 \\ 
SAT-Pro & \textbf{97.9} & \underline{91.1} & 90.9 & 84.4 & 81.6 & 95.1 & 81.4 & 90.1 & \textbf{85.1} & 95.3 & 73.8 & 93.7 & 93.5 & 88.8 \\ 
CT-CLIP & 86.7 & 76.2 & 75.7 & 71.0 & 71.5 & 83.3 & 74.7 & 78.7 & 75.0 & 89.8 & 67.3 & 82.3 & 81.2 & 78.0 \\ 
nnUNet & 97.5 & 92.5 & \underline{93.1} & \underline{85.0} & \underline{84.8} & \underline{98.0} & \underline{83.6} & \underline{90.0} & 82.3 & \underline{95.9} & \textbf{76.6} & \textbf{94.1} & \underline{95.1} & \underline{89.9} \\ 
\textbf{OpenVocabCT} & \underline{97.6} & \textbf{94.3} & \textbf{93.3} & \textbf{86.0} & \textbf{84.9} & \textbf{98.1} & \textbf{83.9} & \textbf{90.1} & \underline{84.2} & \textbf{96.1} & \underline{75.4} & \textbf{94.1} & \textbf{95.7} & \textbf{90.3} \\ \hline
\end{tabular}
\label{tab:flare22_generalization}
\end{table*}

\begin{table*}[ht]
\centering
\caption{Generalization study for the SegTHOR dataset. All models are trained only on TotalSegmentator for fair comparison. Heart$^{*}$ is training unseen organ. Results are reported in DSC (\%). \textbf{BOLD} indicates best result. \underline{Underline} second best result.}
\renewcommand{\arraystretch}{1.2} 
\setlength{\tabcolsep}{8pt} 
\begin{tabular}{l|c|c|c|c|c}
\hline
Method & Esophagus & Heart$^{*}$ & Trachea & Aorta & \textbf{Avg} \\ \hline
Universal & 72.5 & 70.0 & 79.5 & 69.4 & 72.9 \\ 
SAT-Pro & 76.8 & 73.4 & 87.9 & 78.3 & 79.1 \\ 
CT-CLIP & 65.9 & 68.3 & 65.1 & 52.1 & 62.9 \\ 
nnUNet & \underline{83.8} & \underline{78.6} & \textbf{91.3} & \underline{82.4} & \underline{84.0} \\ 
\textbf{OpenVocabCT} & \textbf{85.8} & \textbf{82.8} & \underline{88.9} & \textbf{82.9} & \textbf{85.1} \\ \hline
\end{tabular}
\label{tab:segthor_generalization}
\end{table*}

\section{Conclusion}
In this paper, we propose a novel framework for pretraining and adapting vision-language models for universal text-driven CT image segmentation. Our approach introduces a multi-granular contrastive learning loss that effectively captures organ- and disease-specific information extracted from radiology reports. To ensure high-quality caption selection, we leverage a radiology corpus for generating informative and relevant text descriptions. We first show that our method achieves superior results on organ and lesion segmentation compared to both vision and vision-language models.
We also show that our method can successfully generalize to training-unseen text prompts for universal organ segmentation, outperforming other methods. In our future work, we aim to extend this framework for other imaging modalities (such as PET and MRI) and also explore pretraining with more diverse CT image sites (such as abdominal, head and neck regions). We also plan to transfer our framework to real-world clinical data and demonstrate its effectiveness in other tasks such as tumor detection and image synthesis.

\appendices




\bibliographystyle{IEEEtran} 
\bibliography{tmi} 

\begin{thebibliography}{10}
\providecommand{\url}[1]{#1}
\csname url@samestyle\endcsname
\providecommand{\newblock}{\relax}
\providecommand{\bibinfo}[2]{#2}
\providecommand{\BIBentrySTDinterwordspacing}{\spaceskip=0pt\relax}
\providecommand{\BIBentryALTinterwordstretchfactor}{4}
\providecommand{\BIBentryALTinterwordspacing}{\spaceskip=\fontdimen2\font plus
\BIBentryALTinterwordstretchfactor\fontdimen3\font minus \fontdimen4\font\relax}
\providecommand{\BIBforeignlanguage}[2]{{%
\expandafter\ifx\csname l@#1\endcsname\relax
\typeout{** WARNING: IEEEtran.bst: No hyphenation pattern has been}%
\typeout{** loaded for the language `#1'. Using the pattern for}%
\typeout{** the default language instead.}%
\else
\language=\csname l@#1\endcsname
\fi
#2}}
\providecommand{\BIBdecl}{\relax}
\BIBdecl

\bibitem{ronneberger2015u}
O.~Ronneberger, P.~Fischer, and T.~Brox, ``U-net: Convolutional networks for biomedical image segmentation,'' in \emph{Medical Image Computing and Computer-Assisted Intervention--MICCAI 2015: 18th International Conference, Munich, Germany, October 5-9, 2015, Proceedings, Part III 18}.\hskip 1em plus 0.5em minus 0.4em\relax Springer, 2015, pp. 234--241.

\bibitem{isensee2021nnu}
F.~Isensee, P.~F. Jaeger, S.~A. Kohl, J.~Petersen, and K.~H. Maier-Hein, ``nnu-net: a self-configuring method for deep learning-based biomedical image segmentation,'' \emph{Nature methods}, vol.~18, no.~2, pp. 203--211, 2021.

\bibitem{roy2023mednext}
S.~Roy, G.~Koehler, C.~Ulrich, M.~Baumgartner, J.~Petersen, F.~Isensee, P.~F. Jaeger, and K.~H. Maier-Hein, ``Mednext: transformer-driven scaling of convnets for medical image segmentation,'' in \emph{International Conference on Medical Image Computing and Computer-Assisted Intervention}.\hskip 1em plus 0.5em minus 0.4em\relax Springer, 2023, pp. 405--415.

\bibitem{tang2022self}
Y.~Tang, D.~Yang, W.~Li, H.~R. Roth, B.~Landman, D.~Xu, V.~Nath, and A.~Hatamizadeh, ``Self-supervised pre-training of swin transformers for 3d medical image analysis,'' in \emph{Proceedings of the IEEE/CVF Conference on Computer Vision and Pattern Recognition}, 2022, pp. 20\,730--20\,740.

\bibitem{zhou2023nnformer}
H.-Y. Zhou, J.~Guo, Y.~Zhang, X.~Han, L.~Yu, L.~Wang, and Y.~Yu, ``nnformer: Volumetric medical image segmentation via a 3d transformer,'' \emph{IEEE Transactions on Image Processing}, 2023.

\bibitem{lee20223d}
H.~H. Lee, S.~Bao, Y.~Huo, and B.~A. Landman, ``3d ux-net: A large kernel volumetric convnet modernizing hierarchical transformer for medical image segmentation,'' in \emph{The Eleventh International Conference on Learning Representations}, 2022.

\bibitem{zhao2022prior}
X.~Zhao, P.~Zhang, F.~Song, C.~Ma, G.~Fan, Y.~Sun, Y.~Feng, and G.~Zhang, ``Prior attention network for multi-lesion segmentation in medical images,'' \emph{IEEE Transactions on Medical Imaging}, vol.~41, no.~12, pp. 3812--3823, 2022.

\bibitem{marcus2023concurrent}
A.~Marcus, P.~Bentley, and D.~Rueckert, ``Concurrent ischemic lesion age estimation and segmentation of ct brain using a transformer-based network,'' \emph{IEEE Transactions on Medical Imaging}, vol.~42, no.~12, pp. 3464--3473, 2023.

\bibitem{ji2021learning}
W.~Ji, S.~Yu, J.~Wu, K.~Ma, C.~Bian, Q.~Bi, J.~Li, H.~Liu, L.~Cheng, and Y.~Zheng, ``Learning calibrated medical image segmentation via multi-rater agreement modeling,'' in \emph{Proceedings of the IEEE/CVF Conference on Computer Vision and Pattern Recognition}, 2021, pp. 12\,341--12\,351.

\bibitem{hamamci2024developing}
I.~E. Hamamci, S.~Er, F.~Almas, A.~G. Simsek, S.~N. Esirgun, I.~Dogan, M.~F. Dasdelen, O.~F. Durugol, B.~Wittmann, T.~Amiranashvili \emph{et~al.}, ``Developing generalist foundation models from a multimodal dataset for 3d computed tomography,'' 2024.

\bibitem{moor2023foundation}
M.~Moor, O.~Banerjee, Z.~S.~H. Abad, H.~M. Krumholz, J.~Leskovec, E.~J. Topol, and P.~Rajpurkar, ``Foundation models for generalist medical artificial intelligence,'' \emph{Nature}, vol. 616, no. 7956, pp. 259--265, 2023.

\bibitem{silva2023towards}
J.~Silva-Rodr{\'\i}guez, J.~Dolz, and I.~B. Ayed, ``Towards foundation models and few-shot parameter-efficient fine-tuning for volumetric organ segmentation,'' in \emph{International Conference on Medical Image Computing and Computer-Assisted Intervention}.\hskip 1em plus 0.5em minus 0.4em\relax Springer, 2023, pp. 213--224.

\bibitem{huang2023stu}
Z.~Huang, H.~Wang, Z.~Deng, J.~Ye, Y.~Su, H.~Sun, J.~He, Y.~Gu, L.~Gu, S.~Zhang \emph{et~al.}, ``Stu-net: Scalable and transferable medical image segmentation models empowered by large-scale supervised pre-training,'' \emph{arXiv preprint arXiv:2304.06716}, 2023.

\bibitem{wang2023mis}
G.~Wang, J.~Wu, X.~Luo, X.~Liu, K.~Li, and S.~Zhang, ``Mis-fm: 3d medical image segmentation using foundation models pretrained on a large-scale unannotated dataset,'' \emph{arXiv preprint arXiv:2306.16925}, 2023.

\bibitem{huang2024segment}
Y.~Huang, X.~Yang, L.~Liu, H.~Zhou, A.~Chang, X.~Zhou, R.~Chen, J.~Yu, J.~Chen, C.~Chen \emph{et~al.}, ``Segment anything model for medical images?'' \emph{Medical Image Analysis}, vol.~92, p. 103061, 2024.

\bibitem{zhao2023one}
Z.~Zhao, Y.~Zhang, C.~Wu, X.~Zhang, Y.~Zhang, Y.~Wang, and W.~Xie, ``One model to rule them all: Towards universal segmentation for medical images with text prompts,'' \emph{arXiv preprint arXiv:2312.17183}, 2023.

\bibitem{koleilat2024medclip}
T.~Koleilat, H.~Asgariandehkordi, H.~Rivaz, and Y.~Xiao, ``Medclip-sam: Bridging text and image towards universal medical image segmentation,'' in \emph{International Conference on Medical Image Computing and Computer-Assisted Intervention}.\hskip 1em plus 0.5em minus 0.4em\relax Springer, 2024, pp. 643--653.

\bibitem{kirillov2023segment}
A.~Kirillov, E.~Mintun, N.~Ravi, H.~Mao, C.~Rolland, L.~Gustafson, T.~Xiao, S.~Whitehead, A.~C. Berg, W.-Y. Lo \emph{et~al.}, ``Segment anything,'' in \emph{Proceedings of the IEEE/CVF International Conference on Computer Vision}, 2023, pp. 4015--4026.

\bibitem{liu2023clip}
J.~Liu, Y.~Zhang, J.-N. Chen, J.~Xiao, Y.~Lu, B.~A~Landman, Y.~Yuan, A.~Yuille, Y.~Tang, and Z.~Zhou, ``Clip-driven universal model for organ segmentation and tumor detection,'' in \emph{Proceedings of the IEEE/CVF International Conference on Computer Vision}, 2023, pp. 21\,152--21\,164.

\bibitem{radford2021learning}
A.~Radford, J.~W. Kim, C.~Hallacy, A.~Ramesh, G.~Goh, S.~Agarwal, G.~Sastry, A.~Askell, P.~Mishkin, J.~Clark \emph{et~al.}, ``Learning transferable visual models from natural language supervision,'' in \emph{International conference on machine learning}.\hskip 1em plus 0.5em minus 0.4em\relax PMLR, 2021, pp. 8748--8763.

\bibitem{huang2021gloria}
S.-C. Huang, L.~Shen, M.~P. Lungren, and S.~Yeung, ``Gloria: A multimodal global-local representation learning framework for label-efficient medical image recognition,'' in \emph{Proceedings of the IEEE/CVF International Conference on Computer Vision}, 2021, pp. 3942--3951.

\bibitem{wu2023medklip}
C.~Wu, X.~Zhang, Y.~Zhang, Y.~Wang, and W.~Xie, ``Medklip: Medical knowledge enhanced language-image pre-training for x-ray diagnosis,'' in \emph{Proceedings of the IEEE/CVF International Conference on Computer Vision}, 2023, pp. 21\,372--21\,383.

\bibitem{wang2022medclip}
Z.~Wang, Z.~Wu, D.~Agarwal, and J.~Sun, ``Medclip: Contrastive learning from unpaired medical images and text,'' in \emph{Proceedings of the 2022 Conference on Empirical Methods in Natural Language Processing}, 2022, pp. 3876--3887.

\bibitem{li2023lvit}
Z.~Li, Y.~Li, Q.~Li, P.~Wang, D.~Guo, L.~Lu, D.~Jin, Y.~Zhang, and Q.~Hong, ``Lvit: language meets vision transformer in medical image segmentation,'' \emph{IEEE transactions on medical imaging}, 2023.

\bibitem{pan2023abdomen}
S.~Pan, C.-W. Chang, T.~Wang, J.~Wynne, M.~Hu, Y.~Lei, T.~Liu, P.~Patel, J.~Roper, and X.~Yang, ``Abdomen ct multi-organ segmentation using token-based mlp-mixer,'' \emph{Medical Physics}, vol.~50, no.~5, pp. 3027--3038, 2023.

\bibitem{xue2021multi}
Z.~Xue, P.~Li, L.~Zhang, X.~Lu, G.~Zhu, P.~Shen, S.~A.~A. Shah, and M.~Bennamoun, ``Multi-modal co-learning for liver lesion segmentation on pet-ct images,'' \emph{IEEE Transactions on Medical Imaging}, vol.~40, no.~12, pp. 3531--3542, 2021.

\bibitem{chen2021effective}
C.~Chen, K.~Zhou, M.~Zha, X.~Qu, X.~Guo, H.~Chen, Z.~Wang, and R.~Xiao, ``An effective deep neural network for lung lesions segmentation from covid-19 ct images,'' \emph{IEEE Transactions on Industrial Informatics}, vol.~17, no.~9, pp. 6528--6538, 2021.

\bibitem{xie2022unimiss}
Y.~Xie, J.~Zhang, Y.~Xia, and Q.~Wu, ``Unimiss: Universal medical self-supervised learning via breaking dimensionality barrier,'' in \emph{European Conference on Computer Vision}.\hskip 1em plus 0.5em minus 0.4em\relax Springer, 2022, pp. 558--575.

\bibitem{ma2024segment}
J.~Ma, Y.~He, F.~Li, L.~Han, C.~You, and B.~Wang, ``Segment anything in medical images,'' \emph{Nature Communications}, vol.~15, no.~1, p. 654, 2024.

\bibitem{shi2023generalist}
P.~Shi, J.~Qiu, S.~M.~D. Abaxi, H.~Wei, F.~P.-W. Lo, and W.~Yuan, ``Generalist vision foundation models for medical imaging: A case study of segment anything model on zero-shot medical segmentation,'' \emph{Diagnostics}, vol.~13, no.~11, p. 1947, 2023.

\bibitem{zhang2023biomedclip}
S.~Zhang, Y.~Xu, N.~Usuyama, H.~Xu, J.~Bagga, R.~Tinn, S.~Preston, R.~Rao, M.~Wei, N.~Valluri \emph{et~al.}, ``Biomedclip: a multimodal biomedical foundation model pretrained from fifteen million scientific image-text pairs,'' \emph{arXiv preprint arXiv:2303.00915}, 2023.

\bibitem{liang2023open}
F.~Liang, B.~Wu, X.~Dai, K.~Li, Y.~Zhao, H.~Zhang, P.~Zhang, P.~Vajda, and D.~Marculescu, ``Open-vocabulary semantic segmentation with mask-adapted clip,'' in \emph{Proceedings of the IEEE/CVF Conference on Computer Vision and Pattern Recognition}, 2023, pp. 7061--7070.

\bibitem{ghiasi2022scaling}
G.~Ghiasi, X.~Gu, Y.~Cui, and T.-Y. Lin, ``Scaling open-vocabulary image segmentation with image-level labels,'' in \emph{European Conference on Computer Vision}.\hskip 1em plus 0.5em minus 0.4em\relax Springer, 2022, pp. 540--557.

\bibitem{wang2022cris}
Z.~Wang, Y.~Lu, Q.~Li, X.~Tao, Y.~Guo, M.~Gong, and T.~Liu, ``Cris: Clip-driven referring image segmentation,'' in \emph{Proceedings of the IEEE/CVF conference on computer vision and pattern recognition}, 2022, pp. 11\,686--11\,695.

\bibitem{muller2022radiological}
P.~M{\"u}ller, G.~Kaissis, C.~Zou, and D.~Rueckert, ``Radiological reports improve pre-training for localized imaging tasks on chest x-rays,'' in \emph{International Conference on Medical Image Computing and Computer-Assisted Intervention}.\hskip 1em plus 0.5em minus 0.4em\relax Springer, 2022, pp. 647--657.

\bibitem{zhai2023sigmoid}
X.~Zhai, B.~Mustafa, A.~Kolesnikov, and L.~Beyer, ``Sigmoid loss for language image pre-training,'' in \emph{Proceedings of the IEEE/CVF International Conference on Computer Vision}, 2023, pp. 11\,975--11\,986.

\bibitem{xu2023demystifying}
H.~Xu, S.~Xie, X.~E. Tan, P.-Y. Huang, R.~Howes, V.~Sharma, S.-W. Li, G.~Ghosh, L.~Zettlemoyer, and C.~Feichtenhofer, ``Demystifying clip data,'' \emph{arXiv preprint arXiv:2309.16671}, 2023.

\bibitem{wasserthal2023totalsegmentator}
J.~Wasserthal, H.-C. Breit, M.~T. Meyer, M.~Pradella, D.~Hinck, A.~W. Sauter, T.~Heye, D.~T. Boll, J.~Cyriac, S.~Yang \emph{et~al.}, ``Totalsegmentator: robust segmentation of 104 anatomic structures in ct images,'' \emph{Radiology: Artificial Intelligence}, vol.~5, no.~5, 2023.

\bibitem{antonelli2022medical}
M.~Antonelli, A.~Reinke, S.~Bakas, K.~Farahani, A.~Kopp-Schneider, B.~A. Landman, G.~Litjens, B.~Menze, O.~Ronneberger, R.~M. Summers \emph{et~al.}, ``The medical segmentation decathlon,'' \emph{Nature communications}, vol.~13, no.~1, p. 4128, 2022.

\bibitem{heller2023kits21}
N.~Heller, F.~Isensee, D.~Trofimova, R.~Tejpaul, Z.~Zhao, H.~Chen, L.~Wang, A.~Golts, D.~Khapun, D.~Shats \emph{et~al.}, ``The kits21 challenge: Automatic segmentation of kidneys, renal tumors, and renal cysts in corticomedullary-phase ct,'' \emph{arXiv preprint arXiv:2307.01984}, 2023.

\bibitem{remy2024biolord}
F.~Remy, K.~Demuynck, and T.~Demeester, ``Biolord-2023: semantic textual representations fusing large language models and clinical knowledge graph insights,'' \emph{Journal of the American Medical Informatics Association}, p. ocae029, 2024.

\end{thebibliography}

\end{document}